\documentclass[sigconf]{acmart}

\setcopyright{none}
\renewcommand\footnotetextcopyrightpermission[1]{}

\AtBeginDocument{%
  }

\makeatletter
\def\@authorfont{\large}
\def\@affiliationfont{\small}
\makeatother

\copyrightyear{2026}
\acmYear{2026}
\acmConference[MM '26]
  {The 34th ACM International Conference on Multimedia}
  {November 10--14, 2026}
  {Rio de Janeiro, Brazil}  

\begin{document}

\title{Explainable Deepfake Detection Challenge}


\author{Abhijeet Narang}
\authornote{Equal contribution.}
\email{abhijeet.narang1@monash.edu}
\orcid{0000-0003-0677-910X}
\affiliation{%
  \institution{Monash University}
  \city{Melbourne}
  \country{Australia}
}

\author{Kartik Kuckreja}
\authornotemark[1]
\email{kartikkuckreja456@gmail.com}
\affiliation{%
  \institution{MBZUAI}
  \city{Abu Dhabi}
  \country{UAE}
}

\author{Shreya Ghosh}
\email{shreya.ghosh@uq.edu.au}
\orcid{0000-0002-2639-8374}
\affiliation{%
  \institution{The University of Queensland}
  \city{Brisbane}
  \country{Australia}
}

\author{Muhammad Haris Khan}
\email{muhammad.haris@mbzuai.ac.ae}
\orcid{0000-0001-9746-276X}
\affiliation{%
  \institution{MBZUAI}
  \city{Abu Dhabi}
  \country{UAE}
}

\author{Usman Tariq}
\email{utariq@aus.edu}
\orcid{0000-0002-8244-2165}
\affiliation{%
  \institution{American University of Sharjah}
  \city{Sharjah}
  \country{United Arab Emirates}
  }
  
\author{Jianfei Cai}
\email{jianfei.cai@monash.edu}
\orcid{0000-0002-9444-3763}
\affiliation{%
  \institution{Monash University}
  \city{Melbourne}
  \country{Australia}
}

\author{Abhinav Dhall}
\email{abhinav.dhall@monash.edu}
\orcid{0000-0002-2230-1440}
\affiliation{%
  \institution{Monash University}
  \city{Melbourne}
  \country{Australia}
}

\renewcommand{\shortauthors}{Narang et al.}

\begin{abstract}
Deepfake detection is moving beyond binary classification decisions toward systems that can also explain the visual evidence supporting those decisions. This transition is important for real-world verification settings, where diverse users need to understand not only whether an image is manipulated, but also why it is considered suspicious. The Explainable Deepfake Detection Challenge at ACM Multimedia 2026 is designed to benchmark this joint capability. Built on XPlainVerse, a million-scale benchmark for explainable deepfake detection, the challenge evaluates methods on image classification and grounded natural-language explanation generation. Participants submit a real/fake label together with two explanations for each image: a detailed complex explanation for technical users and a concise simple explanation for general users. The evaluation combines classification metrics with semantic similarity, simplicity, and intent-aware grounding metrics that assess whether explanations identify the relevant manipulated entities and supporting visual evidence. The methodologies developed through the challenge will contribute to the development of next-generation explainable deepfake detectors. Evaluation script, baseline models, and accompanying code are available on \url{https://github.com/Abhijeet8901/XPlainVerse-ACMChallenge}.
\end{abstract}



\maketitle


\section{Introduction}
Recent advances in generative AI have made image manipulation increasingly realistic and difficult to verify, introducing serious risks for digital media security, journalism, and forensic analysis \cite{verdoliva2020media,mirsky2021creation}. While most deepfake detection benchmarks treat this as binary classification: real or fake, this framing is incomplete in practice. A prediction alone rarely satisfies a real verification workflow; analysts, fact-checkers, and moderators need to understand why an image is suspicious, not just that it is.

This motivates explainable deepfake detection as a broader task than classification alone. A useful system should not only identify whether an image has been manipulated, but also communicate the evidence behind that decision in a way that supports trust, verification, and informed action. Such evidence may include inconsistent object boundaries, unnatural shadows, implausible geometry, malformed text, lighting mismatches, or other localized visual artifacts. Critically, effective explanations must go beyond correctness; they should be interpretable, allowing users to understand the reasoning behind a prediction, and usable, enabling them to act on that evidence. Crucially, explanations are not one-size-fits-all: a forensic analyst or technical reviewer may require a precise description of the manipulated region and its specific visual inconsistencies, whereas a general user benefits from a concise summary without specialized terminology. Explanation quality therefore depends not only on whether the evidence is accurate, but on whether it is presented in a form that is meaningful to the intended audience \cite{miller2019explanation,liao2020questioning,sokol2020one}.

The Explainable Deepfake Detection Challenge at ACM Multimedia 2026 builds directly on this need. Grounded in XPlainVerse~\cite{narang2026xplainverse}, a large-scale benchmark pairing authenticity labels with natural-language explanations, the challenge evaluates both classification and explanation generation. For each image, participants produce a real/fake prediction alongside two explanations: one detailed (for technical users) and one simplified (for general audiences). The public release comprises 450K training and 110K validation images, with final rankings on a 200K-image hidden test set. Since detection and explanation are related but distinct capabilities, our metrics span classification performance, semantic similarity, simplicity, and grounding-oriented measures, rewarding systems that are accurate, evidence-grounded, and accessible across user contexts.

\section{Related Work}
Deepfake detection benchmarks have grown steadily in scale and realism, from early datasets like FaceForensics++, DF-TIMIT, and UADFV \cite{rossler2019faceforensics,korshunov2019vulnerability,yang2019exposing} to larger and more diverse collections such as DFDC, Celeb-DF, DeeperForensics, FakeAVCeleb, ForgeryNet and MultiFakeVerse\cite{dolhansky2020deepfake,li2020celebdf,jiang2020deeperforensics,khalid2021fakeavceleb,he2021forgerynet,gupta2025multiverse}. Building on this foundation, the 1M-Deepfakes Detection Challenge introduced large-scale audio-visual deepfake detection and localization as a community benchmark task, while AV-Deepfake1M++ added real-world perturbations \cite{cai2024avdeepfake1m,cai2024challenge,cai2025avdeepfake1mpp}. Our challenge extends this line of work by asking not only whether and where content is manipulated, but whether a system can explain its evidence in language that is meaningful to different users.

Explainable deepfake detection remains relatively underexplored \cite{momin2025explainable}. Visual approaches such as saliency maps, attention overlays, and segmentation masks highlight suspicious regions but do not convey what the evidence means or how to act on it \cite{tsigos2024quantitative}. Recent vision-language methods and benchmarks begin to address this gap through textual or multimodal explanations and broader interpretability evaluation \cite{sun2025visualinguistic,guo2025m2f2,huang2025sida,tariq2025dfp2e,jianglin2026tridf,li2026omnifake}. ExDDV moves toward human-interpretable textual evidence for deepfake video \cite{hondru2026exddv}, and XPlainVerse~\cite{narang2026xplainverse} extends this to images with paired detailed and simplified explanations. This aligns with the broader XAI literature, which emphasizes that explanation quality is inherently user- and context-dependent \cite{miller2019explanation,liao2020questioning,sokol2020one}. Our challenge operationalizes this view, evaluating whether explanations are semantically grounded and usable for both technical reviewers and general audiences.

\section{Challenge Dataset}
\subsection{XPlainVerse Overview}
The challenge uses a curated subset of XPlainVerse~\cite{narang2026xplainverse}, a million-scale benchmark for explainable deepfake detection. It contains both authentic and manipulated images, paired with reference explanations that ground the evaluation of model-generated justifications. The dataset is designed to move beyond face-centric or artifact-specific deepfake detection. Manipulated images cover a diverse range of visual edits, including changes to people, objects, scene layout, local geometry, copied regions, removed or inserted entities, text, and other visually plausible modifications. This diversity encourages systems to reason about image content and visual evidence rather than relying on narrow generator-specific traces.

A key design goal is to support grounded explanation supervision. Manipulated images are generated from caption-guided edit instructions and filtered using an edit-aware quality-control pipeline that compares the original and edited image, identifies visible differences, and verifies that the intended edit is clearly reflected in the final image. This filtering step is critical because explanation supervision becomes unreliable when the requested edit fails or is overwhelmed by off-target artifacts.

The dataset also includes perturbations to evaluate robustness under realistic post-processing and distribution shift, covering common corruptions such as compression, blur, noise, photometric changes, resizing, and frequency-domain transforms, as well as adversarial perturbations. The validation and test splits include more challenging generator families and perturbation conditions to evaluate generalization beyond clean in-distribution samples.

\subsection{Challenge Splits}
Participants receive public training and validation data and submit predictions on a hidden test set. The training split contains 130K real and 320K fake images; the validation split contains 50K real and 60K fake images. The test set of 200K images is used for the final leaderboard.

\begin{table}[t]
\centering
\caption{Challenge split summary. The training and validation splits are released to participants; the hidden test split is used for final ranking.}
\label{tab:split_summary}
\begin{tabular}{lrrr}
\toprule
Split & Real & Fake & Total \\
\midrule
Train & 130,000 & 320,000 & 450,000 \\
Validation & 50,000 & 60,000 & 110,000 \\
Test & 90,000 & 110,000 & 200,000 \\
\bottomrule
\end{tabular}
\end{table}
\section{Challenge Tasks}
The challenge evaluates two complementary capabilities: image-level authenticity classification and natural-language explanation generation. For each test image, participants submit a binary authenticity label together with two explanations targeted at different audiences.

\subsection{Task 1: Image-Level Deepfake Detection}
Given an input image, participants predict whether it is \texttt{real} or \texttt{fake}. Unlike face-centric deepfake benchmarks, the manipulated images in this challenge span a broad range of edit types, including changes to people, objects, scene layout, text, local geometry, and removed or inserted entities. Successful systems should therefore reason about image content broadly rather than relying on face artifacts or generator-specific traces.

\subsection{Task 2: Explanation Generation}

Participants must generate two natural-language explanations for each image, each targeting a different audience. Both explanations should be grounded in visible evidence: for fake images, identifying the suspicious region or entity and the specific visual cues that support the prediction and for real images, identifying natural details consistent with authenticity such as coherent lighting, plausible interactions, and realistic scene structure.

\subsubsection{Complex Explanation}
The complex explanation is intended for technical users such as forensic analysts, researchers, or expert moderators. It should provide a detailed account of the visual evidence, describing concrete cues such as inconsistent boundaries, unnatural shadows, geometry errors, texture failures, lighting mismatch, implausible occlusion, malformed text, or physically unlikely object placement.

\subsubsection{Simple Explanation}
The simple explanation is intended for non-expert users. Rather than enumerating every visual cue, it should communicate the most important evidence in short, clear, accessible language that a general user can quickly act on. This distinction reflects a core design principle of the challenge: that accessibility is itself a dimension of explanation quality, and that no single explanation style is optimal across all verification contexts.
\section{Evaluation Protocol}
\label{sec:evaluation}

Final evaluation is performed on a hidden test set of $N = 200{,}000$ images. The official score combines two components with equal weight: image-level detection performance and explanation quality. For each image $i \in \{1,\ldots,N\}$, let $y_i \in \{0,1\}$ be the ground-truth label, where $0$ denotes real and $1$ denotes fake, and let $\hat{y}_i$ be the submitted prediction.
\vspace{0.5em}

\noindent\textbf{Task 1: Detection.}

The primary detection metric $D_{\mathrm{F1}}$ is macro-F1 over the two authenticity classes:
\begin{equation}
D_{\mathrm{F1}}
=
\frac{1}{2}
\left(
\operatorname{F1}_{\mathrm{fake}}(\hat{\mathbf{y}}, \mathbf{y})
+
\operatorname{F1}_{\mathrm{real}}(\hat{\mathbf{y}}, \mathbf{y})
\right),
\end{equation}
where $\mathbf{y}=\{y_i\}_{i=1}^{N}$ and $\hat{\mathbf{y}}=\{\hat{y}_i\}_{i=1}^{N}$.
\vspace{1em}


\noindent\textbf{Task 2: Explanation.}

For each image, participants submit two natural-language explanations: a complex explanation for technical users and a simple explanation for general users. Let $\hat{e}^{\mathrm{c}}_i$ and $e^{\mathrm{c},\star}_i$ denote the submitted and reference complex explanations for image $i$, and let $\hat{e}^{\mathrm{s}}_i$ and $e^{\mathrm{s},\star}_i$ denote the submitted and reference simple explanations.

The complex explanation score $E_{\mathrm{c}}$ is the mean BERTScore-F1 against the complex references \cite{zhang2020bertscore}:
\begin{equation}
\begin{aligned}
\beta^{\mathrm{c}}_i
&=
\operatorname{BERTScore\text{-}F1}
\bigl(\hat{e}^{\mathrm{c}}_i, e^{\mathrm{c},\star}_i\bigr), \\
B_{\mathrm{c}}
&=
\frac{1}{N}
\sum_{i=1}^{N}
\beta^{\mathrm{c}}_i,
\qquad
E_{\mathrm{c}}
=
B_{\mathrm{c}} .
\end{aligned}
\end{equation}

The simple explanation score $E_{\mathrm{s}}$ combines semantic fidelity and readability. Semantic fidelity is measured by BERTScore-F1 against the simple references:
\begin{equation}
\begin{aligned}
\beta^{\mathrm{s}}_i
&=
\operatorname{BERTScore\text{-}F1}
\bigl(\hat{e}^{\mathrm{s}}_i, e^{\mathrm{s},\star}_i\bigr), \\
B_{\mathrm{s}}
&=
\frac{1}{N}
\sum_{i=1}^{N}
\beta^{\mathrm{s}}_i .
\end{aligned}
\end{equation}

Readability is measured using the Simplicity Level Estimate (SLE) \cite{cripwell2023sle}. Raw SLE values are clipped to $[-1,4]$ and linearly normalized to $[0,1]$. The mean normalized SLE score is:
\begin{equation}
\ell_i
=
\frac{
\operatorname{clip}
\bigl(
\operatorname{SLE}(\hat{e}^{\mathrm{s}}_i), -1, 4
\bigr)
+ 1
}{5},
\qquad
L_{\mathrm{s}}
=
\frac{1}{N}
\sum_{i=1}^{N}
\ell_i .
\end{equation}

The simple explanation score is:
\begin{equation}
E_{\mathrm{s}}
=
0.7\,B_{\mathrm{s}}
+
0.3\,L_{\mathrm{s}} .
\end{equation}

This weighting encourages simple explanations that preserve the meaning of the reference while remaining accessible to non-expert users. The reference-based explanation subscore is the arithmetic mean of the complex and simple explanation scores:
\begin{equation}
E_{\mathrm{ref}}
=
\frac{E_{\mathrm{c}} + E_{\mathrm{s}}}{2}.
\end{equation}

\begin{figure}[t]
    \centering
    \includegraphics[width=0.95\columnwidth]{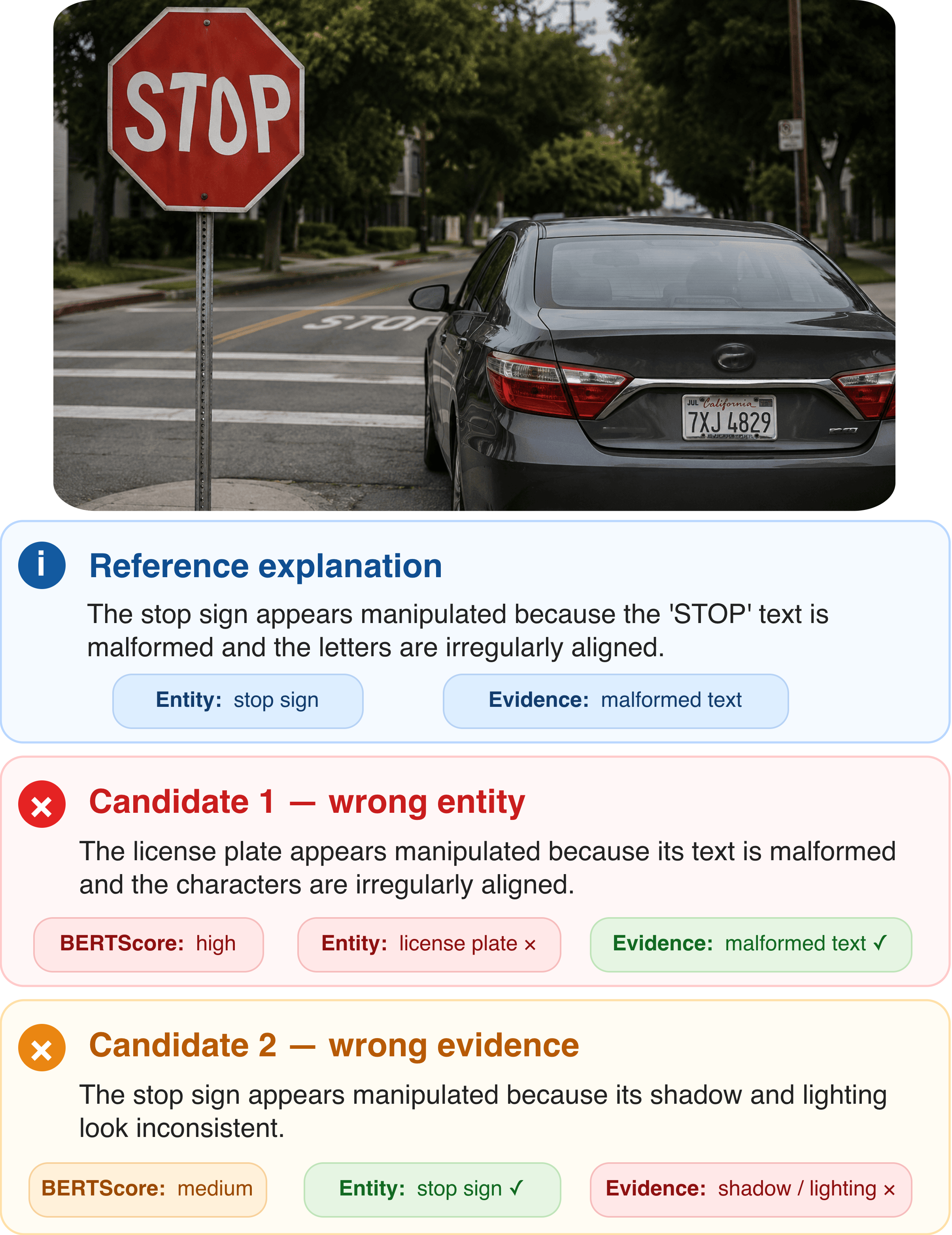}
    \caption{
    Examples where semantic similarity misses grounding errors. Candidate 1 uses evidence wording similar to the reference but attributes it to the wrong entity, while Candidate 2 identifies the correct entity but gives incorrect supporting evidence. These cases show why entity and evidence alignment are needed.
    }
    \label{fig:grounding_motivation}
\end{figure}

\noindent\textbf{LLM-based grounding scores.}

BERTScore measures semantic similarity but does not verify whether an explanation identifies the correct visual evidence. We therefore additionally compute EntityScore and EvidenceScore, introduced in XPlainVerse~\cite{narang2026xplainverse}, using an
LLM evaluator.

For each complex explanation pair, an LLM evaluator extracts diagnostic entities and evidence claims from both the submitted and reference explanations. Let $\Gamma_i^{\mathrm{p}} = (\mathcal{U}_i^{\mathrm{p}}, \mathcal{V}_i^{\mathrm{p}})$ and $\Gamma_i^{\mathrm{r}} = (\mathcal{U}_i^{\mathrm{r}}, \mathcal{V}_i^{\mathrm{r}})$ denote the structures extracted from the submitted and reference explanations respectively, where $\mathcal{U}$ is the set of diagnostic entities, including manipulated or authenticity-relevant objects, subjects, or regions, and $\mathcal{V}$ is the set of evidence claims, including boundary inconsistencies, lighting mismatch, implausible geometry, malformed text, or other visible cues.

To compare the submitted and reference explanations bidirectionally,
we check whether items extracted from one explanation are semantically
supported by the other. Let
$\mathcal Z_i^{q,\mathrm{ent}}=\mathcal U_i^q$ and
$\mathcal Z_i^{q,\mathrm{evid}}=\mathcal V_i^q$, where
$q\in\{\mathrm{p},\mathrm{r}\}$ denotes the submitted or reference
explanation. We define
\begin{equation}
\begin{aligned}
c_i^{\mathrm{p}}(z)
&=\operatorname{cov}
\bigl(z\mid\Gamma_i^{\mathrm{p}},e_i^{\mathrm{c},\star}\bigr),
&
c_i^{\mathrm{r}}(z)
&=\operatorname{cov}
\bigl(z\mid\Gamma_i^{\mathrm{r}},\hat e_i^{\mathrm{c}}\bigr),
\end{aligned}
\label{eq:coverage}
\end{equation}
where each function returns one when the item is semantically
supported by the other explanation, and zero otherwise.

For $m\in\{\mathrm{ent},\mathrm{evid}\}$, precision and recall are
\begin{equation}
P_i^m=
\frac{1}{|\mathcal Z_i^{\mathrm{p},m}|}
\sum_{z\in\mathcal Z_i^{\mathrm{p},m}} c_i^{\mathrm{p}}(z),
\qquad
R_i^m=
\frac{1}{|\mathcal Z_i^{\mathrm{r},m}|}
\sum_{z\in\mathcal Z_i^{\mathrm{r},m}} c_i^{\mathrm{r}}(z).
\label{eq:grounding-pr}
\end{equation}
Precision measures whether the entities or evidence introduced by
the submitted explanation are supported by the reference, while
recall measures whether the reference items are covered by the
submission.

The per-image F1 score and dataset-level grounding score are
\begin{equation}
g_i^m=
\begin{cases}
\dfrac{2P_i^mR_i^m}{P_i^m+R_i^m}, & P_i^m+R_i^m>0,\\
0, & \text{otherwise},
\end{cases}
\qquad
G_m=\frac{1}{N}\sum_{i=1}^{N}g_i^m.
\label{eq:grounding-f1}
\end{equation}
Here, $G_{\mathrm{ent}}$ measures whether explanations identify the
correct entities, while $G_{\mathrm{evid}}$ measures whether they
provide the correct supporting visual evidence.

The grounding, explanation, and final challenge scores are
\begin{equation}
\begin{aligned}
E_{\mathrm{LLM}}
&=\frac{G_{\mathrm{ent}}+G_{\mathrm{evid}}}{2},
&
E_{\mathrm{exp}}
&=0.4E_{\mathrm{ref}}+0.6E_{\mathrm{LLM}},\\
M_{\mathrm{final}}
&=\frac{D_{\mathrm{F1}}+E_{\mathrm{exp}}}{2}.
\end{aligned}
\label{eq:final-score}
\end{equation}

\section{Baselines}
\label{sec:baselines}

We provide baselines as reference points for the evaluation protocol. We fine-tune two open-source vision-language models, Qwen3-VL-8B \cite{bai2025qwen3vl} and InternVL3.5-14B \cite{wang2025internvl35}, on the public training split using LoRA \cite{hu2022lora} for one epoch. Each model is trained with structured prompts that ask it to classify the image as real or fake and generate the requested explanation type, with separate prompts used for complex and simple explanations. Inference is performed at temperature 0.

\begin{table}[!htbp]
\centering
\footnotesize
\setlength{\tabcolsep}{2.8pt}
\caption{Baseline results on the hidden test set. Metrics are defined in Section~5.}
\label{tab:test_baselines}
\vspace{-1em}
\resizebox{\columnwidth}{!}{%
\begin{tabular}{lccccccccc}
\toprule
Model &
$D_{\mathrm{F1}}$ &
$B_{\mathrm{c}}$ &
$B_{\mathrm{s}}$ &
$L_{\mathrm{s}}$ &
$E_{\mathrm{ref}}$ &
$G_{\mathrm{ent}}$ &
$G_{\mathrm{evid}}$ &
$E_{\mathrm{exp}}$ &
$M_{\mathrm{final}}$ \\
\midrule
Qwen3-VL-8B &
0.6342 &
0.6639 &
0.6730 &
0.5575 &
0.6511 &
0.3853 &
0.2583 &
0.4535 &
0.5439 \\
InternVL3.5-14B &
0.6257 &
0.6604 &
0.6695 &
0.5604 &
0.6486 &
0.3586 &
0.2399 &
0.4390 &
0.5323 \\
\bottomrule
\end{tabular}%
}
\vspace{-1em}
\end{table}

\section{Challenge Participation and Results}
\label{sec:challenge-results}

The Explainable Deepfake Detection Challenge 2026 was announced through the official challenge website\footnote{\url{https://explainable-deepfake-detection.github.io/}} and hosted on CodaBench. The challenge received 138 registration-form responses from participating teams and individuals. The public leaderboard was used during the challenge period for automated submission and feedback, while the final results reported here were recomputed by the organizers on the full 200K-image hidden test set using the evaluation protocol described in Section~\ref{sec:evaluation}.

Table~\ref{tab:final_results} presents the official ranking of the five evaluated finalist submissions. To keep the presentation compact, the table reports the final score together with the principal detection, explanation, and grounding metrics; the component-level BERTScore and SLE results are incorporated into $E_{\mathrm{exp}}$ as defined in Section~\ref{sec:evaluation}.

Pixel Sleuth ranked first with a final score of $0.7612$, followed by Team Antvengers ($0.7549$) and MSUteam ($0.7456$). Team Antvengers achieved the highest detection macro-F1 ($0.9479$), whereas Pixel Sleuth obtained the strongest explanation and grounding scores. This ranking reflects the intended design of the benchmark: high detection performance alone is insufficient, as systems must also generate explanations grounded in the correct entities and visual evidence. Team1 further illustrates this distinction, achieving competitive grounding scores despite comparatively lower detection performance.

\begin{table}[!htbp]
\centering
\footnotesize
\setlength{\tabcolsep}{3.8pt}
\caption{Final hidden-test results for the top-five finalist submissions. Metrics are defined in Section~5.}
\label{tab:final_results}
\vspace{-0.5em}
\begin{tabular}{@{}lcccccccc@{}}
\toprule
Team &
$D_{\mathrm{F1}}$ &
$B_{\mathrm{c}}$ &
$B_{\mathrm{s}}$ &
$L_{\mathrm{s}}$ &
$G_{\mathrm{ent}}$ &
$G_{\mathrm{evid}}$ &
$E_{\mathrm{exp}}$ &
$M_{\mathrm{final}}$ \\
\midrule
Pixel Sleuth &
0.9424 & 0.7063 & 0.6819 & 0.9782 & 0.5280 & 0.4205 & 0.5800 & 0.7612 \\
Antvengers &
0.9479 & 0.6958 & 0.6699 & 0.9720 & 0.5078 & 0.3940 & 0.5618 & 0.7549 \\
MSUteam &
0.9340 & 0.7004 & 0.6509 & 0.9726 & 0.4987 & 0.3932 & 0.5571 & 0.7456 \\
Team1 &
0.9167 & 0.6959 & 0.6321 & 0.9335 & 0.5155 & 0.4021 & 0.5590 & 0.7378 \\
HIT VIRLAB &
0.9263 & 0.6988 & 0.6493 & 0.9287 & 0.4467 & 0.3436 & 0.5235 & 0.7249 \\
\bottomrule
\end{tabular}
\vspace{-0.75em}
\end{table}

\noindent\textbf{Participant methods.}
Pixel Sleuth employed a DINOv3-based detection ensemble with contrastive learning and robust augmentation. Its Qwen3-VL explanation model was optimized using preference learning and evidence-grounded rewards to improve both detailed and simplified explanations.

Team Antvengers combined Sapiens2 and DINOv3 detectors through weighted ensembling, while its Qwen3-VL-32B explanation model was trained using supervised fine-tuning followed by GRPO. The method achieved the strongest detection performance but ranked second because of lower explanation-grounding scores.

MSUteam fused features from multiple DINOv3 backbones with a manipulation-localization model, supporting both image-level detection and patch-level artifact reasoning. Separate Qwen3-VL models were used for complex explanation generation and simplification.

Team1 combined an adversarially trained DINOv3 detector with a Qwen3.5-4B explanation model and introduced a reinforcement-learning-based re-examination strategy. HIT VIRLAB proposed a detector-guided framework in which multi-granularity DINOv3 features condition Qwen3-VL-8B to generate detailed explanations that are subsequently simplified.

Overall, DINOv3 was the most common visual backbone, while Qwen-family models were widely adopted for explanation generation. The leading approaches also incorporated detector outputs, spatial evidence, or reward-based optimization rather than relying solely on unconstrained image-to-text generation.

\section{Conclusion and Future Directions}
\label{sec:conclusion}

We presented the Explainable Deepfake Detection Challenge, which jointly evaluates image-level authenticity classification and grounded natural-language explanation generation. Built on XPlainVerse, the challenge requires systems to provide authenticity predictions together with detailed explanations for technical users and simplified explanations for general users. Its evaluation combines detection performance with semantic similarity, readability, and LLM-based entity and evidence grounding.

The results demonstrate that detection and explanation are distinct capabilities: strong classification performance does not necessarily produce explanations grounded in the correct entities and visual evidence. The challenge therefore promotes systems that are not only accurate, but also transparent and useful in practical verification settings. Future work should improve the reliability and efficiency of grounding evaluation, extend coverage to additional manipulations and generation models, support
multilingual explanations, and assess whether explanations help users verify and act on model predictions.

\bibliographystyle{ACM-Reference-Format}
\bibliography{acm_mm_references}

\appendix









\end{document}